\pdfoutput=1

\documentclass[11pt]{article}

\usepackage{EMNLP2022}

\usepackage{times}
\usepackage{latexsym}

\usepackage[T1]{fontenc}

\usepackage[utf8]{inputenc}

\usepackage{microtype}

\usepackage{inconsolata}

\usepackage{array}
\usepackage{pifont}
\usepackage{booktabs}
\usepackage{tabularx}
\usepackage{adjustbox}
\usepackage{multirow}
\usepackage{makecell}
\usepackage{enumitem}
\usepackage{xspace}
\usepackage{tcolorbox}
\usepackage{booktabs,amsfonts,dcolumn}
\usepackage{hyperref}
\usepackage{url}
\usepackage{amsmath,amsthm,amsfonts,amssymb,bm,stmaryrd,bbm}
\newcolumntype{L}[1]{>{\raggedright\let\newline\\\arraybackslash\hspace{0pt}}m{#1}}
\usepackage[ruled, vlined, linesnumbered]{algorithm2e}

\newcommand\tf[1]{\textbf{#1}}

\usepackage{pifont}

\usepackage{CJKutf8}

\title{Retrofitting Multilingual Sentence Embeddings \\ with Abstract Meaning Representation\thanks{~~This work was supported by Alibaba Group through the Alibaba Innovative Research (AIR) Program. $^\dagger$ XL is the corresponding author.}}
\author{Deng Cai$^\heartsuit$\quad Xin Li$^{\spadesuit, \dagger}$\quad Jackie Chun-Sing Ho$^\heartsuit$\quad Lidong Bing$^\spadesuit$\quad Wai Lam$^\heartsuit$\\
	$^\heartsuit$The Chinese University of Hong Kong \\
	$^\spadesuit$DAMO Academy, Alibaba Group \\
	\texttt{thisisjcykcd@gmail.com} \\
	\texttt{\{xinting.lx,l.bing\}@alibaba-inc.com} \\
	\texttt{\{schun,wlam\}@se.cuhk.edu.hk}
	}

\begin{document}
	\maketitle
	\begin{abstract}
	We introduce a new method to improve existing multilingual sentence embeddings with Abstract Meaning Representation (AMR). Compared with the original textual input, AMR is a structured semantic representation that presents the core concepts and relations in a sentence explicitly and unambiguously. It also helps reduce surface variations across different expressions and languages. Unlike most prior work that only evaluates the ability to measure semantic similarity, we present a thorough evaluation of existing multilingual sentence embeddings and our improved versions, which include a collection of five transfer tasks in different downstream applications. Experiment results show that retrofitting multilingual sentence embeddings with AMR leads to better state-of-the-art performance on both semantic textual similarity and transfer tasks. Our codebase and evaluation scripts can be found at \url{https://github.com/jcyk/MSE-AMR}.
	\end{abstract}
	
    \section{Introduction}
    Multilingual sentence embedding (MSE) aims to provide universal sentence representations shared across different languages \citep{hermann-blunsom-2014-multilingual,pham-etal-2015-learning,schwenk-douze-2017-learning}. As an important ingredient of cross-lingual and multilingual natural language processing (NLP), MSE has recently attracted increasing attention in the NLP community. MSE has been widely adopted to bridge the language barrier in several downstream applications such as bitext mining~\citep{guo-etal-2018-effective,schwenk-2018-filtering}, document classification~\citep{eriguchi2018zero,singla-etal-2018-multi,yu-etal-2018-multilingual} and natural language inference~\citep{artetxe-schwenk-2019-massively}. Prior work typically borrows fixed-size embedding vectors from multilingual neural machine models~\citep{schwenk-douze-2017-learning,yu-etal-2018-multilingual} or trains siamese neural networks to align the semantically similar sentences written in different languages~\citep{wieting-etal-2019-simple,yang-etal-2020-multilingual,feng2020language}.
    
    Despite the recent progress, the current evaluation of multilingual sentence embeddings has focused on cross-lingual Semantic Textual Similarity (STS) \cite{agirre-etal-2016-semeval,cer-etal-2017-semeval} or bi-text mining tasks \cite{zweigenbaum2018overview,artetxe-schwenk-2019-massively}. Nevertheless, as pointed out by \newcite{gao2021simcse}, the evaluation on semantic similarity may not be sufficient because better performance on STS does not always indicate better embeddings for downstream tasks. Therefore, for a more comprehensive MSE evaluation, it is necessary to additionally evaluate downstream tasks, which is largely ignored in recent work \cite{chidambaram-etal-2019-learning,reimers-gurevych-2020-making,feng2020language}. In this paper, we collect a set of multilingual transfer tasks and test various existing multilingual sentence embeddings. We find that different methods excel at different tasks and the conclusions drawn from the STS evaluation do not always hold in the transfer tasks and vice versa. We aim to establish a standardized evaluation protocol for future research in multilingual sentence embeddings.

    To improve the quality of existing MSE models, we explore Abstract Meaning Representation (AMR) \cite{banarescu-etal-2013-abstract}, a symbolic semantic representation, for augmenting existing neural semantic representations. Our motivation is two-fold. First, AMR explicitly offers core concepts and relations in a sentence. This helps prevent learning the superficial patterns or spurious correlations in the training data, which do not generalize well to new domains or tasks~\citep{poliak-etal-2018-hypothesis, clark-etal-2019-dont}. Second, AMR reduces the variances in surface forms with the same meaning. This helps alleviate the data sparsity issue as there are rich lexical variations across different languages.
	 
	On the other hand, despite that AMR is advocated to act as an interlingua~\cite{xue-etal-2014-interlingua,hajic-etal-2014-comparing,damonte-cohen-2018-cross}, little work has been done to reflect on the ability of AMR to have impact on subsequent tasks. In order to advance research in AMR and its applications, multilingual sentence embedding can be seen as an important benchmark for highlighting its ability to abstract away from surface realizations and represent the core concepts expressed in the sentence. To our knowledge, this is the first attempt to leverage the AMR semantic representation for multilingual NLP.
	
	We learn AMR embeddings with contrastive siamese network \cite{gao2021simcse} and AMR graphs derived from different languages \cite{cai2021multilingual}. Experiment results on 10 STS tasks and 5 transfer tasks with four state-of-the-art embedding methods show that retrofitting multilingual sentence embeddings with AMR improves the performance substantially and consistently.
	
	Our contribution is three-fold.
	\begin{itemize}[wide=0\parindent,noitemsep,topsep=0em]
		\item We propose a new method to obtain high-quality semantic vectors for multilingual sentence representation, which takes advantage of language-invariant Abstract Meaning Representation that captures the core semantics of sentences.
		\item We present a thorough evaluation of multilingual sentence embeddings, which goes beyond semantic textual similarity and includes various transfer tasks in downstream applications.
		\item We demonstrate that retrofitting multilingual sentence embeddings with Abstract Meaning Representation leads to better performance on both semantic textual similarity and transfer tasks.
	\end{itemize}
	\section{Related Work}
	\paragraph{Universal Sentence Embeddings}
	Our work aims to learn universal sentence representations, which should be useful for a broad set of applications. There are two lines of research for universal sentence embeddings: unsupervised approaches and supervised approaches. Early unsupervised approaches \cite{kiros2015skip-thought,hill-etal-2016-learning,gan-etal-2017-learning,logeswaran2018an-quick-thought} design various surrounding sentence reconstruction/prediction objectives for sentence representation learning. \newcite{jernite2017discourse} exploit sentence-level discourse relations as supervision signals for training sentence embedding model. Instead of using the interactions of sentences within a document, \newcite{le2014distributed} propose to learn the embeddings for texts of arbitrary length on top of word vectors. Likewise, \newcite{chen2017efficient,pagliardini-etal-2018-unsupervised,yang-etal-2019-parameter} calculate sentence embeddings from compositional $n$-gram features. Recent approaches often adopt contrastive objectives \cite{zhang-etal-2020-unsupervised,giorgi2021declutr,wu2020clear,meng2021coco,carlsson2021semantic,kim2021self,yan2021consert,gao2021simcse} by taking different views---from data augmentation or different copies of models---of the same sentence as training examples.
	
	On the other hand, supervised methods \cite{conneau-etal-2017-supervised-infersent,cer-etal-2018-universal,reimers-gurevych-2019-sentence,gao2021simcse} take advantage of labeled natural language inference (NLI) datasets \cite{bowman-etal-2015-large-snli,williams-etal-2018-broad-mnli}, where a sentence embedding model is fine-tuned on entailment or contradiction sentence pairs. Furthermore, \citet{wieting-gimpel-2018-paranmt,wieting-etal-2020-bilingual} demonstrate that bilingual and back-translation corpora provide useful supervision for learning semantic similarity. Another line of work focuses on regularizing embeddings~\cite{li-etal-2020-sentence,su2021whitening,huang2021whiteningbert} to alleviate the representation degeneration problem.
	\paragraph{Multilingual Sentence Embeddings}
	Recently, multilingual sentence representations have attracted increasing attention. \newcite{schwenk-douze-2017-learning,yu-etal-2018-multilingual,artetxe-schwenk-2019-massively} propose to use encoders from multilingual neural machine translation to produce universal representations across different languages. \newcite{chidambaram-etal-2019-learning,wieting-etal-2019-simple,yang-etal-2020-multilingual,feng2020language} fine-tune siamese networks \cite{bromley1993signature} with contrastive objectives using parallel corpora. \newcite{reimers-gurevych-2020-making} train a multilingual model to map sentences to the same embedding space of an existing English model. Different from existing work, our work resorts to multilingual AMR, a language-agnostic disambiguated semantic representation, for performance enhancement.
	\paragraph{Evaluation of Sentence Embeddings}
	Traditionally, the mainstream evaluation for assessing the quality of \textit{English-only} sentence embeddings is based on the Semantic Textual Similarity (STS) tasks and a suite of downstream classification tasks. The STS tasks \citep{agirre-etal-2012-semeval,agirre-etal-2013-sem,agirre-etal-2014-semeval,agirre-etal-2015-semeval,agirre-etal-2016-semeval,marelli-etal-2014-sick,cer-etal-2017-semeval} calculate the embedding distance of sentence pairs and compare them with the human-annotated scores for semantic similarity. The classification tasks (e.g., sentiment analysis) from SentEval \cite{conneau-kiela-2018-senteval} take sentence embeddings as fixed input features to a logistic regression classifier. These tasks are commonly used to benchmark the transferability of sentence embeddings on downstream tasks. For \textit{multilingual} sentence embeddings, most previous work has focused on cross-lingual STS \cite{agirre-etal-2016-semeval,cer-etal-2017-semeval} and the relevant bi-text mining tasks \cite{zweigenbaum2018overview,artetxe-schwenk-2019-massively}. The evaluation on downstream transfer tasks has been largely ignored \cite{chidambaram-etal-2019-learning,reimers-gurevych-2020-making,feng2020language}. Nevertheless, as pointed out in \newcite{gao2021simcse} in English scenarios, better performance on semantic similarity tasks does not always indicate better embeddings for transfer tasks. For a more comprehensive evaluation, in this paper, we collect a set of multilingual transfer tasks and test various existing multilingual sentence embeddings. We aim to establish a standardized evaluation protocol for future research in multilingual sentence embeddings.
	
	\section{Preliminaries}
	\subsection{Contrastive Siamese Network}
	\label{siamese}
	Siamese network \cite{bromley1993signature} has attracted considerable attention for self-supervised representation learning. It has been extensively adopted with contrastive learning \cite{hadsell2006dimensionality} for learning dense vector representations of images and sentences \cite{reimers-gurevych-2019-sentence,chen2020simple}. The core idea of contrastive learning is to pull together the representations of semantically close objects (images or sentences) and repulse the representations of negative pairs of dissimilar ones. Recent work in computer vision \cite{caron2020unsupervised,grill2020bootstrap,chen2021exploring,zbontar2021barlow} has demonstrated that negative samples may not be necessary. A similar observation was made in NLP by \newcite{zhang-etal-2021-bootstrapped} who adopted the BYOL framework \cite{grill2020bootstrap} for sentence representation learning. In this work, we adopt the framework in \cite{gao2021simcse} with in-batch negatives \cite{chen2017sampling,henderson2017efficient}. Formally, we assume a set of training examples $\mathcal{D}=\{(x_i, x_i^{+}, x_i^{-})\}_{i=1}^N$, where $x_i^{+}$ and $x_i^{-}$ are semantically close and semantically irrelevant to $x_i$, respectively. The training is done with stochastic mini-batches. Each mini-batch consists of $M$ examples and the training objective is defined as:
	\begin{equation}
	\ell_i = - \log \frac{e^{s(x_i,x_i^{+})/\tau}}{\sum_{j=1}^{M}e^{s(x_i,x_j^{-})/\tau} + \sum_{j=1}^{M}e^{s(x_i,x_j^{
				+})/\tau}}
	\label{loss}
	\end{equation}
	where $s(\cdot,\cdot)$ measures the similarity of two objects and $\tau$ is a scalar controlling the temperature of training. As seen, other objects in the same mini-batch (i.e., $\{x_j^{-}\}_{j\ne i}$ and $\{x_j^{+}\}_{j\ne i}$) are treated as negatives for $x_i$. More concretely, $s(\cdot,\cdot)$ computes the cosine similarity between the representations of two objects:
	\begin{equation}
	s(x_i, x_j) = \frac{\mathbf{h}_i^{\texttt{T}} \mathbf{h}_j}{\|\mathbf{h}_i\| \cdot\|\mathbf{h}_j\|}
	\nonumber
	\end{equation}
	where $\mathbf{h}_i$ and $\mathbf{h}_j$ are obtained from a neural encoder $f_{\theta}(\cdot)$: $\mathbf{h} =  f_{\theta}(x)$. The model parameters $\theta$ are then optimized using the contrastive learning objective.
	\subsection{Multilingual AMR Parsing}
	\label{parsing}
	AMR \cite{banarescu-etal-2013-abstract} is a broad-coverage semantic formalism originally designed for English. The accuracy of AMR parsing has been greatly improved in recent years \cite{cai-lam-2019-core,cai-lam-2020-amr,bevilacqua2021one,bai-etal-2022-graph}. Because AMR is agnostic to syntactic and wording variations, recent work has suggested the potential of AMR to work as an \textit{interlingua} \cite{xue-etal-2014-interlingua,hajic-etal-2014-comparing,damonte-cohen-2018-cross}. That is, we can represent the semantics in other languages using the corresponding AMR graph of the semantic equivalent in English. A number of \textit{cross-lingual} AMR parsers \cite{damonte-cohen-2018-cross,blloshmi-etal-2020-xl,sheth2021bootstrapping,procopio2021sgl,cai2021multilingual} have been developed to transform non-English texts into AMR graphs. Most of them rely on pre-trained multilingual language models and synthetic parallel data. In particular, \newcite{cai2021multilingual} proposed to learn a multilingual AMR parser from an English AMR parser via knowledge distillation. Their single parser is trained for five different languages (German, Spanish, Italian, Chinese, and English) and achieves state-of-the-art parsing accuracies. In addition, the one-for-all design maintains parsing efficiency and reduces prediction inconsistency across different languages. Thus, we adopt the multilingual AMR parser of \newcite{cai2021multilingual} in our experiments.\footnote{\url{https://github.com/jcyk/XAMR}}
	
	It is worth noting that the multilingual parser is capable of parsing many other languages, including those it has not been explicitly trained for, thanks to the generalization power inherited from pre-trained multilingual language models \cite{tang2020multilingual,liu2020multilingual}. In Section \ref{incorprate-parsing}, we further extend the training of the multilingual parser to French, another major language, for improved performance.
	\section{Proposed Method}
	We first introduce how we learn AMR embeddings and then describe the whole pipeline for enhancing existing sentence embeddings.
	\subsection{Learning AMR Embeddings}
	
	\paragraph{Linearization \& Modeling}
	Given AMR is graph-structured, a variety of graph neural networks \cite{song-etal-2018-graph,beck-etal-2018-graph,ribeiro-etal-2019-enhancing,guo2019densely,cai2020graph,ribeiro-etal-2019-enhancing} have been proposed for the representation learning of AMR. However, recent work \cite{zhang-etal-2019-amr,mager-etal-2020-gpt,bevilacqua2021one} has demonstrated that the power of existing pre-trained language models based on the Transformer architecture \cite{vaswani2017attention}, such as BERT \cite{devlin-etal-2019-bert}, GPT2 \cite{radford2019language} and BART \cite{lewis-etal-2020-bart}, can be leveraged for achieving better performance. Following them, we also take BERT as the backbone model.

\begin{figure}
		\includegraphics[width=0.98\linewidth]{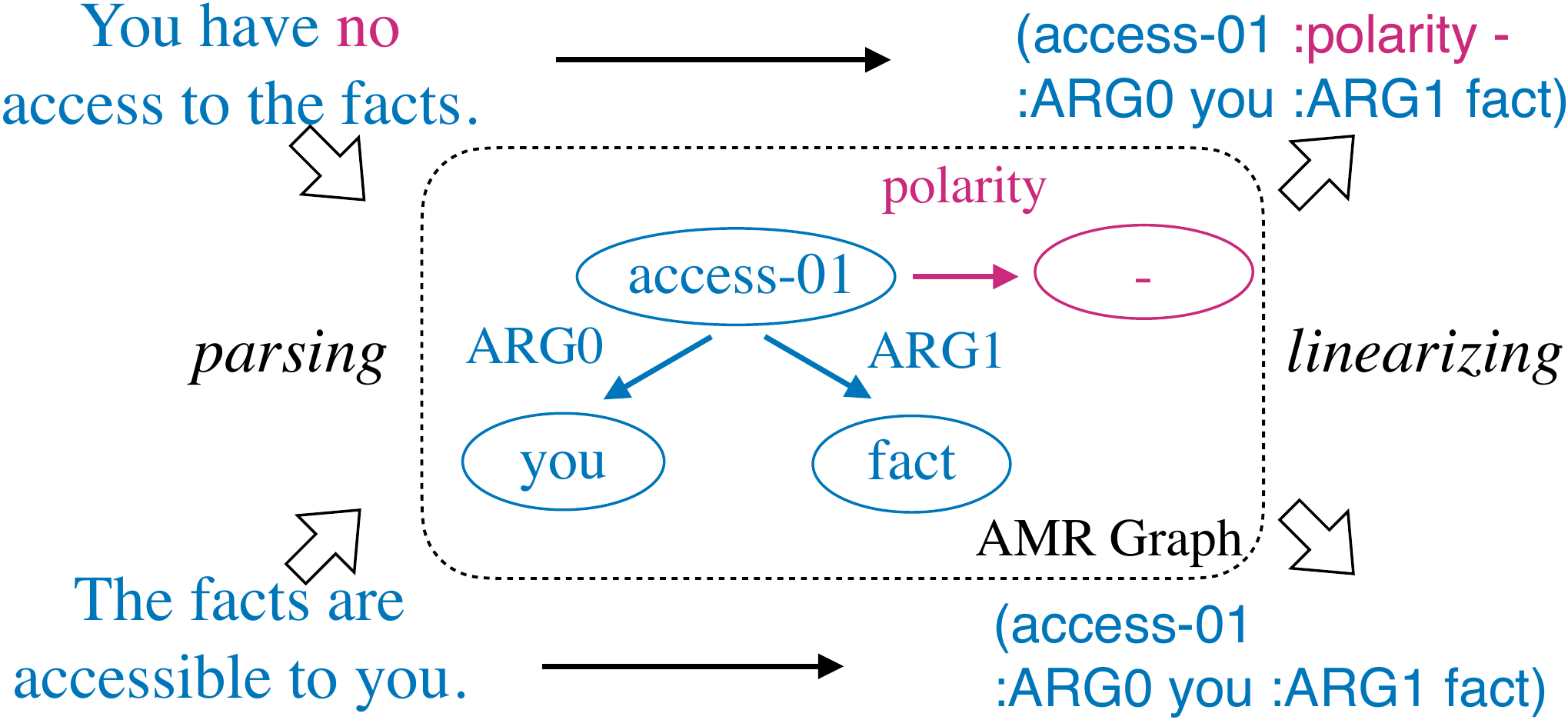}
		\caption{The parsing and linearization pipeline.}
		\label{fig}
	\end{figure}

	Since Transformer-based language models are designed for sequential data, to encode graphical AMR, we resort to the linearization techniques in \cite{bevilacqua2021one}. Figure \ref{fig} illustrates the linearization of AMR graphs. For each AMR graph, a DFS traversal is performed starting from the root node of the graph, and the trajectory is recorded. We use parentheses to mark the hierarchy of node depths. \newcite{bevilacqua2021one} also proposed to use special tokens for indicating variables in the linearized graph and for handling reentrancies (i.e., a node plays multiple roles in the graph). However, the introduction of special tokens significantly increases the length of the output sequence (almost 50\% increase). We remove this feature and simply repeat the nodes when re-visiting happens. This significantly reduces the length of the output sequence and allows more efficient modeling with Transformer-based language models. The downside is that reentrancy information becomes unrecoverable. However, we empirically found that the shortened sequences lead to better performance. The linearizations of AMR graphs are then treated as plain token sequences when being fed into Transformer-based language models.
	Note that AMR linearization introduces additional tokens that are rarely shown in English (e.g., ``ARG2" and ``belong-01"). These tokens may not be included in the original vocabulary of existing language models and could be segmented into sub-tokens  (e.g., ``belong-01" $\Rightarrow$ ``belong", ``-", ``01"), which are less meaningful and increase the sequence length. To deal with this problem, we extend the original vocabulary of existing language models to include all the relation and frame names occurring at least 5 times in the AMR sembank (LDC2017T10).
	\paragraph{Positive \& Negative Examples}
	Contrastive learning aims to learn effective representations by pulling semantically similar examples together and pushing apart dissimilar examples. Following the discussion in Section \ref{siamese}, the most critical question in contrastive learning is how to obtain positive and negative examples. In language representations, positive examples $x_i^{+}$ are often constructed by applying minimal distortions (e.g., word deletion, reordering, and substitution) on $x_i$ \cite{wu2020clear,meng2021coco} or introducing some random noise (e.g., dropout \cite{srivastava2014dropout}) to the modeling function $f_{\theta}$ \cite{gao2021simcse}. On the other hand, negative examples $x_i^{-}$ are usually sampled from other sentences. However, prior work \cite{conneau-etal-2017-supervised-infersent,gao2021simcse} has demonstrated that entailment/contradiction sentence pairs in supervised natural language inference (NLI) datasets \cite{bowman-etal-2015-large-snli,williams-etal-2018-broad-mnli} are better positive/negative pairs for learning sentence embeddings. Following \cite{gao2021simcse}, we borrow the supervisions from two NLI datasets, namely SNLI \cite{bowman-etal-2015-large-snli} and MNLI \cite{williams-etal-2018-broad-mnli}. In the NLI datasets, given one premise, there are one entailment hypothesis and another contradiction hypothesis accompanying. Therefore, in each training example $(x_i, x_i^{+}, x_i^{-})$, $x_i$ is the premise, $x_i^{+}$ is the entailment hypothesis, and $x_i^{-}$ is the contradiction hypothesis.
	
	Specifically, we use the multilingual AMR parser described in Section \ref{parsing} to parse sentences into AMR graphs. Because the sentences in the NLI datasets are in English, the resultant AMR graphs are all derived from English. This is in contrast to downstream applications where an AMR graph may be derived from a foreign language. To reduce the discrepancy between training and testing, we use OPUS-MT \cite{tiedemann-thottingal-2020-opus} \footnote{\url{https://huggingface.co/docs/transformers/model_doc/marian}}, an off-the-shelf translation system, to translate English sentences in the NLI datasets to other languages. The translations in other languages are then parsed by our multilingual AMR parser. In this way, we extend the training of AMR embeddings to multilingual scenarios as well.
	\paragraph{Mixed Training} To better cover both the monolingual and cross-lingual settings in downstream applications, the training aims to capture the interactions between AMR graphs derived from the same language as well as those derived from different languages. To this end, we mix up AMR graphs from different languages during training. Moreover, to alleviate the drawback of imperfect parsing and avoid catastrophic forgetting of pre-trained language models, we also mix up AMR graphs and original English sentences during training. The details are shown in Algorithm \ref{sampling}.
	
	We hypothesize that the noise introduced by automatic translation could negatively affect the performance but a suitable amount of noise might also serve as a helpful regularizer. Unfortunately, due to the lack of gold translations, we could not perform a rigorous quantitative comparison. In our preliminary experiments, we also tried another automatic translation system, mBART-mmt \cite{tang2020multilingual} \footnote{\url{https://huggingface.co/facebook/mbart-large-50-many-to-many-mmt}}, other than OPUS-MT. We found that mBART-mmt leads to worse performance in general, likely due to its lower translation quality.
	
	\subsection{Incorporating AMR Embeddings}
	The learned AMR embeddings can be used to augment any existing sentence embedding model. For any input sentence $x$, it is processed through two channels: (1) the sentence is first parsed into an AMR graph $y=\texttt{parse}(x)$. The graph is then fed into our AMR encoder: $\mathbf{h}=f_\theta(y)$. (2) the sentence is directly encoded by an off-the-shelf sentence embedding model $g(\cdot)$: $\mathbf{s}=g(x)$. Lastly, we combine the text and graph embeddings ($\mathbf{s}$ and $\mathbf{h}$) to produce the final sentence representation. 
	\paragraph{Parsing}
	\label{incorprate-parsing}
	Theoretically, the multilingual AMR parser introduced in \newcite{cai2021multilingual} can parse 50 different languages as it inherits the multilingual encoder pre-trained on these languages from \newcite{tang2020multilingual}. However, the original parser has only been explicitly trained for German (de), Spanish (es), Italian (it), Chinese (zh), and English (en). We hypothesize that including more languages in training can help improve the overall parsing accuracy. Therefore, we add French (fr), another major language, to the training of the parser.\footnote{The extension only requires an English-to-French translation system, which is the OPUS-MT system in our implementation. We refer readers to \newcite{cai2021multilingual} for more details.} 
	\paragraph{Integration}
	We explore four different choices for the integration of the text embedding $\mathbf{s}$ and the AMR embedding $\mathbf{h}$: $\mathbf{s} \oplus \mathbf{h}$, $\mathbf{s} + \mathbf{h}$, ,$ \frac{\mathbf{s}}{\|\mathbf{s}\|} \oplus \frac{\mathbf{h}}{\|\mathbf{h}\|}$ ,$\frac{\mathbf{s}}{\|\mathbf{s}\|} + \frac{\mathbf{h}}{\|\mathbf{h}\|}$, where $\oplus$ denotes the concatenation of two vectors. Empirically, we find that $ \frac{\mathbf{s}}{\|\mathbf{s}\|} \oplus \frac{\mathbf{h}}{\|\mathbf{h}\|}$ generally works best.
			\begin{algorithm}[t]
	\small
	\DontPrintSemicolon
	\SetInd{0.1em}{0.8em}
	\KwIn{Dataset: $\mathcal{D}=\{(x_i, x_i^{+}, x_i^{-})\}_{i=1}^N$, \\ Systems: AMR parser $\texttt{parse}(\cdot)$ and English-to-$l$ translator $\texttt{translate}(\cdot, l)$, Maximum training steps: $T$, Batch size $M$, Language set: $\mathcal{L}$.}
			\For{$t \gets 1$ to $T$}{                 
			    Draw a mini-batch $\mathcal{B}=\{(x_i, x_i^{+}, x_i^{-})\}_{i=1}^M$ from $\mathcal{D}$\;
			    \ForEach{ \textnormal{sentence} x \textnormal{\textbf{in}} $\mathcal{B}$}{
			       Draw a language $l \sim \mathcal{L}$\;
			        \uIf {$l$ \textnormal{\textbf{is not}} \text{en}}{
			            $x \gets \texttt{parse}(\texttt{translate}(x, l)$)\;
			        }
			        \Else{
			        Draw a text/graph factor $q \sim U(0, 1)$\;
			            \If {$q$ > 0.5}{
			                $x \gets \texttt{parse}(x)$\;
			            }
			            
			        }
			    }
			Optimize the model $f_\theta$ with Eq. (\ref{loss}) on the updated $\mathcal{B}$\;}  
			\KwOut {Optimized Model $f_\theta$}
		\caption{Learning AMR Embeddings.}
		\label{sampling}
	\end{algorithm}
	\section{Evaluation Benchmark}
	To provide a more comprehensive evaluation of multilingual sentence representations, In addition to traditional semantic textual similarity tasks, we also introduce a set of downstream transfer tasks.
	\subsection{Semantic Textual Similarity}
	\paragraph{Multilingual STS}
	The goal of semantic textual similarity (STS) is to assign for a pair of sentences a score indicating their semantic similarity. For example, a score of 0 indicates not related and 5 indicates semantically equivalent. We use the datasets in \newcite{reimers-gurevych-2020-making}, which is an extended version of the multilingual STS 2017 dataset \cite{cer-etal-2017-semeval}. The evaluation is done by comparing the distance in the embedding space and the human-annotated scores in the dataset. 
	
	\subsection{Transfer Tasks}
	We evaluate the quality of the multilingual sentence embeddings on the following cross-lingual sentence/sentence-pair classification benchmarks:
	\paragraph{XNLI} The Cross-lingual Natural Language Inference benchmark~\cite{conneau-etal-2018-xnli} is used to estimate the capability of cross-lingual / multilingual models in recognizing textual entailment. The evaluation sets of XNLI are created by manually translating the development corpus and the testing corpus of MultiNLI~\cite{williams-etal-2018-broad-mnli} to 15 other languages. 
	
	\paragraph{PAWS-X} The Cross-lingual Paraphrase Adversaries from Word Scrambling benchmark~\cite{yang-etal-2019-paws} consists of golden English paraphrase identification pairs from PAWS~\cite{zhang-etal-2019-paws} and around 24k human translations of PAWS evaluation sets (i.e., development set and testing set) in English, French, Spanish, German, Chinese, Japanese (ja), and Korean (ko).
	
	\paragraph{QAM} The Question-Answer Matching task aims to predict if the given (question, passage) pair is a QA pair. We use the multilingual QAM dataset from XGLUE~\cite{liang-etal-2020-xglue}, which provides the labeled instance (question, passage, label) in English, French, and German, to evaluate the effectiveness of multilingual sentence embeddings.
	
	\paragraph{MLDoc} The Multilingual Document Classification benchmark~\cite{schwenk-li-2018-corpus} is a multilingual corpus with a collection of news documents written in English, German, Spanish, French, Italian, Chinese, Japanese, and Russian (ru). The entire corpus is manually classified into four groups according to the topic of the document.
	\paragraph{MARC} The Multilingual Amazon Review Corpus~\cite{keung-etal-2020-multilingual} is a large-scale collection of Amazon user reviews for multilingual rating classification. The corpus covers 6 languages, including English, German, French, Spanish, and Chinese, Japanese.
	\section{Experiments}
	\subsection{Experimental Setup}
	For STS tasks, following previous work \cite{gao2021simcse}, we define the similarity score as the cosine similarity of sentence embeddings and compute the Spearman’s rank correlation between the computed score and the gold score.
	
	For downstream transfer tasks, we follow the conventional zero-shot cross-lingual transfer setting \cite{liang-etal-2020-xglue,hu2020xtreme}, where annotated training data is provided in English but none is provided in other languages. We fit a logistic regression classifier on top of fixed sentence representations and follow default configurations in \newcite{conneau-kiela-2018-senteval,gao2021simcse}. To faithfully reflect the multilinguality of multilingual sentence embeddings, we train exactly one model for each task. The union of the development sets in different languages is adopted for model selection.
		\begin{table}[t]
		\centering
		\small
		\begin{tabular}{lL{5.3cm}}
			\Xhline{3\arrayrulewidth}
			Task & Languages \\  \hline
			XNLI & en, fr, de ,es, zh, el, bg, ru, tr, ar, vi, th, hi, sw, ur \\
			PAWS-X & en, fr, de, es, ja, ko \\
			QAM & en, fr, de \\
			MLDoc & en, fr, de, es, zh, ru, it, ja \\
			MARC & en, fr, de, es, zh, ja \\
			\Xhline{3\arrayrulewidth}
		\end{tabular}
		\caption{Test languages in different transfer tasks.}
		\label{tab:downstream_task_info}
	\end{table}
	\subsection{Implementation Details}
	\label{sec:impl}
	We initialize our AMR encoder with BERT \cite{devlin-etal-2019-bert} (uncased) and take the \texttt{[CLS]} representation as the sentence embedding. By default, the AMR encoder is trained on English, German, Spanish, Italian, Chinese, French, and Arabic (ar).  Each model is trained for a maximum of $9$ epochs with a learning rate of $5e-5$ and a batch size of $512$. The temperature in Eq. (\ref{loss}) is set to be $0.05$. For model selection, we use the STS-B development \cite{cer-etal-2017-semeval}. We train a multilingual AMR parser on English, German, Spanish, Italian, Chinese, and French using the same recipe in \newcite{cai2021multilingual}. We will release our code and models.
	\subsection{Baseline Systems}
	We evaluate the following systems:
	\paragraph{mBERT / XLM-R} We use the mean pooling of the outputs from the pre-trained mBERT \cite{devlin-etal-2019-bert} and XLM-R \cite{conneau2020unsupervised}, which are pre-trained on multilingual data. However, no parallel or labeled data was used.
	\paragraph{mUSE} Multilingual Universal Sentence Encoder \cite{chidambaram-etal-2019-learning} uses a dual-encoder transformer architecture and adopts contrastive objectives. It was trained on mined question-answer pairs, SNLI data, translated SNLI data, and parallel corpora over 16 languages.
	\paragraph{LASER} LASER \cite{artetxe-schwenk-2019-massively} trains a sequence-to-sequence encoder-decoder architecture on parallel corpora for machine translation. The sentence representation is obtained via max-pooling over the output of the encoder. LASER was trained over 93 languages.
	\paragraph{LaBSE} Language-agnostic BERT Sentence Embedding (LaBSE) \cite{feng2020language} was trained similar to mUSE with a translation ranking loss. It fine-tunes a dual-BERT architecture with 6 Billion translation pairs for 109 languages.
	\paragraph{Xpara} \newcite{reimers-gurevych-2020-making} fine-tunes XLM-R to imitate SBERT-paraphrases \cite{reimers-gurevych-2019-sentence}, a RoBERTa model trained on more than 50 Million English paraphrase pairs, with massive bilingual sentence pairs over 50 languages.
	\subsection{Model Variants}
	
	To study the effect of each modeling choice, we implement a series of model variants.
	\begin{itemize}[wide=0\parindent,noitemsep,topsep=0em]
	\item \textbf{\#1}: To show if learning from English data suffices, we train the AMR encoder with only English sentences and the AMR graphs derived from them.
	\item \textbf{\#2}: To study the effect of extending the training of the multilingual AMR parser to French, we use the original parser in \newcite{cai2021multilingual}, which does not include French.
	\item \textbf{\#3}: To measure the help of involving more languages when training the AMR encoder, we train the AMR encoder without the AMR graphs derived from French and Arabic.
	\item \textbf{\#4}: To validate the usefulness of adding the English sentences to the training of the AMR encoder, we train the AMR encoder without English sentences.
	\item \textbf{\#5}: The standard model as described in Section \ref{sec:impl}.
	\end{itemize}
	For each model variant, we report the average performance over five different runs (different random seeds) throughout this paper.
	\subsection{Results}
			\begin{table}[t]
		\centering
		\small
		\setlength{\tabcolsep}{4pt}
		\begin{tabular}{lrcccl}
			\toprule
			\textbf{Model} & \textbf{\#}& \textbf{EN-EN} & \textbf{ES-ES} & \textbf{AR-AR}   & \textbf{Avg.} ($\Delta$) \\
			\midrule
			mBERT && 54.36 & 56.69    &    50.86    & 53.97 \\
            XLM-R && 52.18    &    49.58    &    25.50    & 42.42 \\
            mUSE  && 86.39    &    \underline{86.86}    &    76.41    & \underline{83.22} \\
            LASER && 77.62   &    79.69    &    68.84    & 75.38 \\
            LaBSE && 79.45    &    80.83    &    69.07    & 76.45 \\
            Xpara && \underline{88.10}    &    85.71    &    \underline{79.10}    & \underline{84.30} \\
            \hline
\multirow{5}{*}{mUSE++}&1&88.51&86.53&80.12&85.05 (+1.83)\\
&2&88.57&\textbf{87.57}&80.45&85.53 (+2.31)\\
&3&88.30&87.07&80.32&85.23 (+2.01)\\
&4&88.38&86.95&80.56&85.30 (+2.08)\\
&5&88.74&87.14&80.67&85.52 (+2.30)\\
\hline
\multirow{5}{*}{Xpara++}&1&89.31&85.89&80.62&85.28 (+0.98)\\
&2&89.19&86.60&\textbf{81.85}&\textbf{85.88 (+1.58)}\\
&3&89.06&86.40&80.78&85.42 (+1.12)\\
&4&89.27&86.34&80.74&85.45 (+1.15)\\
&5&\textbf{89.45}&86.52&81.04&85.66 (+1.36)\\
			\bottomrule
		\end{tabular}
		\caption{
			Performance (Spearman’s correlation) on STS tasks (monolingual setup). $\Delta$ indicates the improvements from our methods.
		}
		\label{tab:sts-mono}
	\end{table}
    \begin{table*}[t]
		\centering
		\small
		\begin{tabular}{lrcccccccl}
			\toprule
			\textbf{Model} &\textbf{\#}& \textbf{EN-AR} & \textbf{EN-DE} & \textbf{EN-TR} & \textbf{EN-ES} & \textbf{EN-FR} & \textbf{EN-IT} & \textbf{EN-NL} & \textbf{Avg.} ($\Delta$)\\
			\midrule
			mBERT && 18.67   &    33.86    &    16.02    &    21.47    &    32.98    &    34.02    &    35.30    & 27.47  \\
XLM-R &&  15.71 & 21.30    &    12.07    &    10.60    &    16.64    &    22.88    &    23.95    & 17.59\\
mUSE  && 79.27    &    82.13    &    75.47    &    79.62    &    82.64    &    84.55    &    \underline{84.07}    & \underline{81.11} \\
LASER && 66.53    &    64.20    &    71.99    &    57.93    &    69.06    &    70.83    &    68.67    & 67.03 \\
LaBSE && 74.51    &    73.85    &    72.07    &    65.71    &   76.98    &    76.99    &    75.22    & 73.62\\
Xpara && \underline{81.81}    &    \underline{83.66}    &    \underline{80.16}    &    \underline{84.05}    &    \underline{83.16}    &    \underline{85.66}    &    83.67    & \underline{83.17}\\
\hline
\multirow{5}{*}{mUSE++}&1&79.99&83.81&75.08&81.74&83.39&86.84&86.61&82.49 (+1.38)\\
&2&81.85&85.01&75.12&83.25&83.68&85.62&84.78&82.76 (+1.65)\\
&3&80.22&84.18&76.53&82.79&84.60&86.75&86.28&83.05 (+1.94)\\
&4&80.43&84.41&76.58&82.59&84.52&86.72&86.38&83.09 (+1.98)\\
&5&80.52&84.87&76.57&83.01&84.91&86.71&\textbf{86.71}&83.33 (+2.22)\\
\hline
\multirow{5}{*}{Xpara++}&1&81.16&84.86&77.75&83.71&83.61&87.32&85.42&83.40 (+0.23)\\
&2&\textbf{82.89}&85.56&77.66&\textbf{85.14}&84.44&86.35&84.08&83.73 (+0.56)\\
&3&81.47&85.28&79.21&84.55&84.77&87.02&85.15&83.92 (+0.75)\\
&4&81.45&85.58&79.20&84.47&84.84&87.13&85.34&84.00 (+0.83)\\
&5&81.73&\textbf{85.62}&\textbf{79.50}&84.76&\textbf{85.22}&\textbf{87.33}&85.58&\textbf{84.25 (+1.08)}\\

			\bottomrule
		\end{tabular}
		
		\caption{
			Performance (Spearman’s correlation) on STS tasks (cross-lingual setup).
		}
		\label{tab:sts-cross}
	\end{table*}
		\begin{table*}[t]
		\centering
		\small
		\begin{tabular}{lrcccccl}
			\toprule
			\multirow{2}{*}{\tf{Model}} &
			 \multirow{2}{*}{\tf{\#}} & \tf{MLDoc} & \tf{XNLI} & \tf{PAWS-X} & \tf{MARC} & \tf{QAM} & \tf{Avg.} (\tf{$\Delta$}) \\
			  && seen/all & seen/all & seen/all & seen/all  &   & seen/all  \\
			\midrule
			mBERT && 80.17/77.90 & 47.23/44.41 & 57.30/57.05 & 38.66/38.43 & 55.25 & 55.72/54.61 \\
			XLM-R && 79.99/77.86 & 48.57/46.83 & 56.10/56.06 & 50.63/50.03 & 55.58 & 58.17/57.27 \\
			mUSE  && 79.79/77.20 & 55.60/48.63 & 57.68/57.34 & 47.28/46.37 & \underline{60.82} & 60.23/58.07 \\
			LASER && 77.42/74.63 & \underline{60.36}/\underline{58.87} & \underline{73.89}/\underline{70.81} & 49.08/47.97 & 58.28 & \underline{63.81/62.11} \\
			LaBSE && \underline{84.93}/\underline{82.29} & 58.24/56.65 & 58.75/58.31 & \underline{49.95}/\underline{48.85} & 59.34 & \underline{62.24/61.09} \\
	        Xpara && 65.68/62.42 & 55.81/53.33 & 58.50/58.06 & 49.92/48.79 & 56.25 & 57.23/55.77 \\
	        \hline
\multirow{5}{*}{LASER++}&1&81.67/78.67&63.48/57.53&73.64/70.49&49.42/48.32&59.16&65.47/62.83 (+1.66/+0.72)\\
&2&81.71/78.71&63.64/57.33&73.51/70.36&49.46/48.18&59.08&65.48/62.73 (+1.67/+0.62)\\
&3&81.91/79.03&63.65/57.86&73.68/70.50&49.62/48.51&59.37&65.65/63.05 (+1.84/+0.94)\\
&4&82.74/79.74&63.45/57.66&73.72/70.52&49.32/48.27&59.42&65.73/63.12 (+1.92/+1.01)\\
&5&82.65/79.80&\textbf{63.88}/\textbf{57.98}&\textbf{73.79}/\textbf{70.61}&49.44/48.31&59.41&\textbf{65.83}/\textbf{63.22} (\textbf{+2.02}/\textbf{+1.11})\\
\hline
\multirow{5}{*}{LaBSE++}&1&85.59/82.86&59.24/53.06&59.92/59.13&51.08/50.07&59.73&63.11/60.97 (+0.87/-0.12)\\
&2&85.68/82.77&59.66/53.07&59.44/58.80&50.87/49.84&59.85&63.10/60.87 (+0.86/-0.22)\\
&3&85.56/82.82&59.69/53.54&59.66/58.95&51.15/50.08&59.99&63.21/61.08 (+0.97/-0.01)\\
&4&\textbf{85.89}/\textbf{83.10}&59.44/53.31&59.68/58.98&51.13/50.11&\textbf{60.21}&63.27/61.14 (+1.03/+0.05)\\
&5&85.70/83.02&59.66/53.55&59.81/59.07&\textbf{51.20}/\textbf{50.21}&59.99&63.27/61.17 (+1.03/+0.08)\\

			\bottomrule
		\end{tabular}
		
		\caption{
			Performance (accuracy) on transfer tasks. $\Delta$ indicates the improvements from our methods.
		}
		\label{tab:ml_transfer}
	\end{table*}
	\paragraph{Multilingual STS}
    Table \ref{tab:sts-mono} and Table \ref{tab:sts-cross} show the evaluation results on 3 monolingual STS tasks and 7 cross-lingual STS tasks respectively. As seen, the best-performing models in the literature are mUSE and Xpara. Thus, we present the results of augmenting mUSE and Xpara with our AMR embeddings, denoted by mUSE++ and Xpara++ respectively. Using AMR embeddings substantially improves both two models across the monolingual (up to +2.31 on avg.) and cross-lingual settings (up to +2.22 on avg.), greatly advancing the state-of-the-art performance. The average scores of monolingual and cross-lingual settings are pushed to 85.88 and 84.25 respectively. The improvements for mUSE are generally greater than those for Xpara, even though the training data of mUSE overlaps with our AMR encoders. We hypothesize that it is because Xpara is trained on paraphrase corpus, which diminishes the ability of AMR to group different expressions of the same semantics.
    
    One interesting finding is that model variant \#2 performs best on monolingual settings while model variant \#5 attains the best results on cross-lingual settings. We believe that adding more languages to the training of the AMR parser helps the generalization to other languages and reduces the parsing inconsistency across different languages. Thus, the AMR graphs from different languages are better aligned, leading to a better-aligned vector space. On the other hand, adding more language may decrease the parsing accuracies on individual languages due to the fixed model capacity. Note that all other model variants except \#2 underperform \#5, confirming the effectiveness of the proposed mixed training strategy.
	\paragraph{Transfer Tasks}
	Table \ref{tab:ml_transfer} shows the evaluation results on transfer tasks. For each task, we report the macro-average scores across different languages. The results for each language can be found in Appendix. Different to previous work, our AMR encoders are only trained with a few languages (en, de, es, it, zh, fr, and ar) at most. To isolate the effect on unseen languages, we separate the results on those seen languages from all languages (seen/all). First of all, we find that the rankings of existing models are quite different to the results on STS tasks. LASER and LaBSE achieve the best results on most transfer tasks except for QAM, and outperforms mUSE and Xpara by large margins in most cases. The results demonstrate the limitation of solely testing on sentence similarity measurement.
	
	Next, we augment the best-performing models, LASER and LaBSE, with our AMR embeddings (LASER++ and LaBSE++). For seen languages, our methods substantially boost the performance of these two models across different tasks (up to +2.02 on avg.). The performance gains over LASER are greater than those over LaBSE. Note that LASER is trained with an encoder-decoder architecture and both LaBSE and our AMR encoders are trained with a Siamese network. Therefore, we believe the AMR embeddings are more complementary to LASER.
	
	When considering all languages, the improvements over LASER are also considerable (up to +1.11 on avg.). However, according to the average scores over different tasks, the AMR embeddings seem to fail to improve LaBSE; We even observe a performance drop for model variants \#1-\#3. Nevertheless, the performance drop largely comes from XNLI while the scores on other tasks are instead boosted. This is because the test sets of XNLI include some distant languages (e.g., Swahili and Urdu) that our multilingual AMR parser cannot handle well (see the results on individual languages in Table \ref{tab:xnli} in Appendix). We conjecture that further extending the multilingual AMR parser to more languages can alleviate this problem. The comparison among different model variants provides a basis for the above speculation. As we can see, model variant \#2 (exclude French from the training of the multilingual AMR parser) performs worst. Also, model variants \#1 (drop all non-English AMR graphs for training) and \#2 (drop the AMR graphs derived from French and Arabic) are the other two variants that negatively impact the average performance. Another interesting observation is that model variant \#4 performs best on MLDoc and QAM, suggesting English sentences might not be necessary.
	\section{Conclusion}
	This paper presented a thorough evaluation of existing multilingual sentence embeddings, ranging from traditional text similarity measurement to a new variety of transfer tasks. We found that different methods excel at different tasks. We then proposed to improve existing methods with universal AMR embeddings, which leads to better performance on all tasks.
	
	\section*{Limitations}
	Although our work provides an effective solution for improving multilingual sentence embeddings with AMR, we acknowledge some limitations of this study and further discuss them in the following: (1) Our framework treats the text encoder as a black box and does not care too much about its implementation. Although it is flexible and straightforward to apply our framework to any multilingual sentence embedding model, designing more specific interaction mechanisms for different text encoders is supposed to be better and we leave it as future work. (2) The improvement from our framework is higher in seen languages than unseen languages. Further extending the language coverage in the training phases of both the multilingual AMR parser and the AMR encoder is presumably beneficial to the cross-lingual generalization capability of the AMR embeddings. However, due to the limit of computational resources, we only consider a few languages in the experiments.
	\bibliography{anthology,custom}
	\bibliographystyle{acl_natbib}
	\appendix
\section{Transfer task}
We provide the detailed results on each language for each transfer task in Table \ref{tab:mldoc}-\ref{tab:qam}.

\begin{table*}[ht]
    \centering
    \small
    \begin{tabular}{lrccccccccc}
    \toprule
       \tf{Model} & \tf{\#}& \tf{DE}  &   \tf{EN}  &   \tf{ES}  &   \tf{FR}  &   \tf{IT}  &   \tf{JA}  &   \tf{RU}  &   \tf{ZH} & \tf{Avg.}\\
           \midrule
            mBERT && 83.73 & 89.88 & 75.75 & 83.73 & 68.25 & 71.12 & 71.08 & 79.65 & 77.90 \\
            XLM-R && 84.60 & 88.78 & 77.98 & 80.20 & 74.00 & 73.25 & 69.70 & 74.38 & 77.86 \\
            mUSE && 85.80 & 87.95 & 77.00 & 84.45 & 68.45 & 69.35 & 69.50 & 75.08 & 77.20 \\
            LASER && 84.28 & 84.95 & 72.85 & 79.25 & 69.67 & 67.10 & 65.42 & 73.50 & 74.63 \\
            LaBSE && 88.42 & 90.88 & 81.03 & 87.90 & 76.40 & 73.00 & 75.70 & 84.97 & 82.29 \\
            Xpara && 69.23 & 88.35 & 67.88 & 65.15 & 61.62 & 52.45 & 52.83 & 41.85 & 62.42 \\
            \hline
           \multirow{5}{*}{LASER++}&1&87.99&89.22&79.41&82.75&71.33&72.72&66.62&79.35&78.67\\
&2&87.56&89.80&79.35&82.65&71.77&71.64&67.78&79.16&78.71\\
&3&87.62&89.49&79.65&83.02&71.29&73.69&67.08&80.39&79.03\\
&4&88.22&90.27&80.39&83.93&72.17&73.99&67.53&81.45&79.74\\
&5&88.02&90.10&80.17&83.59&72.54&74.38&68.12&81.50&79.80\\

\hline
\multirow{5}{*}{LaBSE++}&1&89.38&91.70&83.28&88.07&76.22&75.45&73.87&84.87&82.86\\
&2&89.72&91.64&83.72&87.84&76.41&74.73&73.32&84.76&82.77\\
&3&89.08&91.46&83.59&88.07&76.08&75.53&73.67&85.09&82.82\\
&4&89.41&91.45&84.51&88.30&76.37&75.75&73.76&85.27&83.11\\
&5&89.39&91.41&84.11&88.00&76.28&75.67&74.30&85.02&83.02\\



    \bottomrule
    \end{tabular}
    
    \caption{
        MLDoc results of different sentence embedding models.
    }
    \label{tab:mldoc}
\end{table*}
\begin{table*}[ht]
    \centering
    \scriptsize
    \begin{tabular}{lrcccccccccccccccc}
    \toprule
       \tf{Model} & \tf{\#} & \tf{AR}  &\tf{BG}  &\tf{DE}  &\tf{EL}  &\tf{EN}  &\tf{ES}  &\tf{FR}  &\tf{HI}  &\tf{RU}  &\tf{SW}  &\tf{TH}  &\tf{TR}  &\tf{UR}  &\tf{VI}  &\tf{ZH} & \tf{Avg.}\\
           \midrule
            mBERT && 42.57 & 45.35 & 46.75 & 43.99 & 53.53 & 47.64 & 47.60 & 41.54 & 46.07 & 37.49 & 36.75 & 43.17 & 40.46 & 47.96 & 45.31 & 44.41 \\
            XLM-R && 45.15 & 48.90 & 45.85 & 49.28 & 54.53 & 49.56 & 49.96 & 45.23 & 47.35 & 38.64 & 42.28 & 47.43 & 41.92 & 49.94 & 46.37 & 46.83 \\
            mUSE && 53.09 & 48.76 & 55.05 & 35.39 & 59.02 & 56.25 & 55.81 & 35.91 & 54.93 & 39.20 & 54.23 & 53.59 & 37.86 & 35.97 & 54.35 & 48.63 \\
            LASER && 58.80 & 60.26 & 60.96 & 60.68 & 61.54 & 60.60 & 60.52 & 56.19 & 59.50 & 53.13 & 59.26 & 59.46 & 52.46 & 59.92 & 59.76 & 58.87 \\
            LaBSE && 56.75 & 57.56 & 57.07 & 57.88 & 60.58 & 58.74 & 58.08 & 55.57 & 56.41 & 54.77 & 53.79 & 56.43 & 52.08 & 55.89 & 58.22 & 56.65 \\
            Xpara && 54.33 & 56.15 & 56.45 & 55.97 & 57.07 & 55.95 & 56.31 & 52.18 & 55.31 & 34.17 & 53.67 & 54.73 & 48.20 & 54.75 & 54.75 & 53.33 \\
            \hline
            \multirow{5}{*}{LASER++}&1&58.52&51.44&63.94&45.89&67.37&63.83&65.02&57.12&61.74&43.72&49.46&59.67&51.40&61.58&62.20&57.53\\
&2&58.10&48.34&64.45&45.63&67.59&65.03&64.17&57.32&61.58&44.29&49.98&58.57&51.34&60.98&62.52&57.33\\
&3&58.76&51.51&64.45&46.45&67.26&64.46&64.96&57.16&62.44&43.94&50.04&60.29&51.84&62.36&62.01&57.86\\
&4&58.67&51.59&64.02&46.39&66.76&64.24&64.71&57.09&62.09&43.88&49.73&60.08&51.51&61.90&62.30&57.67\\
&5&59.10&51.66&64.84&46.25&67.02&64.61&65.19&57.66&62.80&44.00&49.89&60.23&51.78&62.07&62.55&57.98\\

\hline
\multirow{5}{*}{LaBSE++}&1&54.90&45.94&58.92&39.92&63.98&59.72&60.04&54.54&56.88&39.56&42.32&55.40&49.04&56.87&57.85&53.06\\
&2&55.38&43.67&59.54&40.08&64.28&60.88&59.24&54.47&56.81&40.22&41.74&54.76&49.47&56.95&58.62&53.07\\
&3&55.98&46.60&59.68&40.21&63.93&60.70&59.99&54.79&57.50&39.64&42.42&56.24&49.90&57.61&57.85&53.54\\
&4&55.91&46.34&59.30&40.71&63.49&60.52&59.87&54.55&57.18&39.38&42.09&56.09&49.47&57.24&57.56&53.31\\
&5&56.41&46.39&59.30&40.67&63.63&60.51&60.16&55.12&57.31&39.73&42.26&56.21&49.80&57.77&57.93&53.55\\

    \bottomrule
    \end{tabular}
    \caption{
        XNLI results of different sentence embedding models.
    }
    \label{tab:xnli}
\end{table*}
\begin{table*}[ht]
    \centering
    \small
    \begin{tabular}{lrcccccccc}
    \toprule
       \tf{Model} & \tf{\#}& \tf{DE}  & \tf{EN}  & \tf{ES}  & \tf{FR} & \tf{JA} & \tf{KO} & \tf{ZH}  & \tf{Avg.}\\
           \midrule
            mBERT && 57.00 & 57.30 & 57.45 & 57.40 & 56.85 & 56.00 & 57.35 & 57.05 \\
            XLM-R && 55.70 & 55.70 & 55.65 & 55.25 & 56.05 & 55.85 & 58.20 & 56.06 \\
            mUSE && 57.70 & 58.10 & 56.45 & 57.35 & 56.70 & 56.25 & 58.80 & 57.34 \\
            LASER && 72.20 & 79.80 & 75.00 & 74.80 & 65.40 & 60.85 & 67.65 & 70.81  \\
            LaBSE && 58.80 & 58.90 & 57.55 & 59.50 & 57.10 & 57.30 & 59.00 & 58.31  \\
            Xpara && 59.00 & 57.35 & 58.35 & 59.30 & 57.45 & 56.50 & 58.50 & 58.06 \\
            \hline
            \multirow{5}{*}{LASER++}&1&72.13&80.06&74.62&74.34&64.41&60.79&67.05&70.48\\
&2&72.51&79.57&74.67&74.12&64.49&60.46&66.69&70.36\\
&3&72.15&80.34&74.59&74.26&64.37&60.75&67.06&70.50\\
&4&72.09&80.18&74.92&74.52&64.26&60.79&66.87&70.52\\
&5&72.23&80.01&74.82&74.74&64.37&60.91&67.16&70.61\\

\hline
\multirow{5}{*}{LaBSE++}&1&59.50&61.61&58.74&60.28&57.25&57.06&59.49&59.13\\
&2&59.17&60.87&57.95&59.16&57.51&56.88&60.04&58.79\\
&3&59.83&60.69&58.43&59.70&57.30&57.01&59.66&58.95\\
&4&59.42&60.63&58.77&60.01&57.40&57.03&59.58&58.98\\
&5&59.55&61.25&58.61&60.05&57.29&57.13&59.61&59.07\\

    \bottomrule
    \end{tabular}
    \caption{
        PAWS-X results of different sentence embedding models.
    }
    \label{tab:paws}
\end{table*}
%
%
%

\begin{table*}[ht]
    \centering
    \small
    \begin{tabular}{lrccccccc}
    \toprule
       \tf{Model} & \tf{\#}&\tf{DE}  &   \tf{EN} & \tf{ES} & \tf{FR}  &   \tf{JA} & \tf{ZH} & \tf{Avg.}\\
           \midrule
            mBERT && 38.28 & 45.54 & 38.32 & 38.40 & 32.78 & 37.28 & 38.43 \\
            XLM-R && 52.16 & 54.78 & 48.70 & 48.08 & 49.42 & 47.02 & 50.03 \\
            mUSE && 47.18 & 50.90 & 48.52 & 47.76 & 42.02 & 41.82 & 46.37 \\
            LASER && 51.44 & 52.68 & 49.26 & 50.00 & 42.02 & 42.42 & 47.97 \\
            LaBSE && 51.46 & 52.40 & 49.86 & 50.46 & 45.58 & 43.36 & 48.85 \\
            Xpara && 52.24 & 53.50 & 49.22 & 50.12 & 44.50 & 43.14 & 48.79 \\
            \hline
            \multirow{5}{*}{LASER++}&1&51.48&53.90&50.22&49.92&41.58&42.84&48.32\\
&2&51.76&54.02&50.25&49.90&41.39&41.76&48.18\\
&3&51.54&54.24&50.18&50.23&41.90&42.97&48.51\\
&4&51.74&54.13&49.64&50.04&41.04&43.06&48.28\\
&5&51.70&54.20&49.80&50.42&41.07&42.68&48.31\\
\hline
\multirow{5}{*}{LaBSE++}&1&53.26&54.10&50.78&51.07&46.18&45.01&50.07\\
&2&52.80&54.24&50.78&50.45&46.06&44.69&49.84\\
&3&53.27&54.35&50.71&51.07&46.34&44.74&50.08\\
&4&53.33&54.02&50.92&51.16&46.23&45.03&50.12\\
&5&53.53&54.33&50.77&51.34&46.02&45.29&50.21\\


    \bottomrule
    \end{tabular}
    \caption{
        MARC results of different sentence embedding models.
    }
    \label{tab:marc}
\end{table*}

\begin{table*}[ht]
    \centering
    \small
    \begin{tabular}{lrcccc}
    \toprule
       \tf{Model} & \tf{\#}&\tf{DE}  &   \tf{EN} & \tf{FR} & \tf{Avg.}\\
           \midrule
            mBERT && 54.21 & 56.60 & 54.94 & 55.25\\
            XLM-R && 55.30 & 57.18 & 54.25 & 55.58  \\
            mUSE && 62.60 & 58.01 & 61.84 & 60.82  \\
            LASER && 57.95 & 58.63 & 58.25 & 58.28 \\
            LaBSE && 59.06 & 58.15 & 60.82 & 59.34 \\
            Xpara && 58.90 & 57.01 & 60.08 & 58.66 \\
            \hline
            \multirow{5}{*}{LASER++}&1&59.17&59.62&58.69&59.16\\
&2&58.81&59.52&58.91&59.08\\
&3&59.24&59.96&58.89&59.37\\
&4&59.46&59.88&58.91&59.42\\

&5&59.34&59.94&58.94&59.41\\
\hline
\multirow{5}{*}{LaBSE++}&1&61.03&57.42&60.75&59.73\\
&2&60.81&57.26&61.48&59.85\\
&3&60.94&57.99&61.05&59.99\\
&4&61.23&58.48&60.91&60.21\\
&5&60.55&58.83&60.60&59.99\\

    \bottomrule
    \end{tabular}
    \caption{
        QAM results of different sentence embedding models.
    }
    \label{tab:qam}
\end{table*}
\end{document}